\documentclass[cameraready]{Interspeech}
\usepackage{tcolorbox}
\usepackage{booktabs,multirow}
\usepackage{float}
\usepackage{caption}

\title{A Finetuned SpeechLLM for Joint Multi-Granular L2 Assessment and Natural-Language Rationales}

\author[orcid=0009-0001-9244-9491]{Aditya Kamlesh}{Parikh}
\author[orcid=0000-0001-5395-0438]{Cristian}{Tejedor-Garcia}
\author[orcid=0000-0001-5908-0824]{Catia}{Cucchiarini}
\author[orcid=0000-0003-1722-3465]{Helmer}{Strik}

\address{
    $^1$ Centre for Language Studies, Radboud University, Nijmegen, The Netherlands}

\email{aditya.parikh@ru.nl, cristian.tejedorgarcia@ru.nl, catia.cucchiarini@ru.nl,
helmer.strik@ru.nl}


\usepackage{comment}

\begin{document}

\maketitle

\begin{abstract}
Automated L2 speech assessment can assign proficiency labels, but often lacks interpretability. We propose a rubric-guided SpeechLLM for multi-aspect, multi-granular assessment, trained with a hybrid objective combining supervised fine-tuning and Bounded Direct Preference Optimization. The model jointly predicts ordinal labels at the sentence-level (accuracy, fluency, prosody), word/phoneme-level accuracy, and generates a natural-language rationale in the same response. On SpeechOcean762, our approach matches or outperforms single-granularity models while remaining competitive with prior approaches. We analyze rationale reliability along two axes: self-consistency with model predictions and alignment with ground-truth labels, using sentiment consistency (plausibility) and mention-based agreement (faithfulness). Rationales are plausible at the sentence level, but faithfulness degrades at the word/phoneme level: references are sparse and weakly aligned with token-level labels.
    
\end{abstract}

\section{Introduction}
The growing demand for effective second language (L2) acquisition has intensified interest in instructional approaches that target oral proficiency. Despite advances in pedagogy and digital learning environments, spoken communication remains one of the most challenging competencies for L2 learners to develop \cite{derakhshan2015interference}. Difficulties in speech production stem from multiple sources, including first-language (L1) phonological transfer among adult learners and developmental constraints shaping younger learners' phonetic and phonological acquisition \cite{baker2008child}. While overcoming these barriers requires intensive, individualized feedback, delivering such support through one-on-one instruction is not scalable. Computer-Assisted Language Learning (CALL) systems therefore offer a low-pressure alternative that enables learners to engage in autonomous oral practice \cite{dong2022fostering}. To be effective, they must provide interpretable, multi-granular assessment and diagnostic information about phoneme and word accuracy, prosody, fluency \cite{saito2021effects}, and intelligibility \cite{derwing2018putting}.

Early approaches \cite{witt2000phone,cucchiarini2000quantitative,cucchiarini2009oral} and modern self-supervised models (wav2vec~2.0 \cite{baevski2020wav2vec}, HuBERT \cite{9585401}) have advanced mispronunciation detection and intelligibility scoring \cite{xu2021explore,kim22k_interspeech,parikh25_interspeech,cao24b_interspeech,parikh25_interspeech_logit}; however, they typically output opaque numeric metrics or token-level labels that lack actionable pedagogical insight. Although subsequent frameworks introduced multi-granular evaluation across sentence, word, and phoneme levels \cite{lin20f_interspeech,9746743,chao23_interspeech}, generating coherent natural-language rationales that are consistent with assigned labels and informative across granularities remains a critical, unresolved gap.

Large Language Models (LLMs) have expanded the scope of interpretable spoken language assessment by enabling natural-language rationales alongside predicted scores \cite{fayyaz2024evaluating}. However, for L2 speech assessment, such rationales should be supported by acoustic evidence \cite{chen2025read}. Standard cascaded ASR--LLM pipelines \cite{wang2025exploring} can produce fluent rationales, but they do not directly leverage the acoustic signal and therefore cannot reliably ground assessments in speech evidence. This has motivated end-to-end (E2E) SpeechLLMs that integrate acoustic and linguistic representations \cite{wu2023decoder,tang2023salmonn,chu2023qwen}, and more recently instruction-tuned SpeechLLMs that better follow complex natural-language instructions and generate more reliable, structured responses \cite{chu2024qwen2,ghosh-etal-2024-gama,pmlr-v267-ghosh25b}. Building on these instruction-tuned models, recent work has started to study explanatory speech assessment setups, where a model outputs both an evaluative judgment and a free-text rationale \cite{wang2025speechllm}. Nevertheless, the application of such models to rubric-guided L2 speech assessment remains limited, highlighting both promise and open challenges. For instance, \cite{parikh2025zero} found that rubric-based SpeechLLMs exhibit a ``niceness bias'' in zero-shot settings. Supervised fine-tuning (SFT) can mitigate this bias \cite{ma25b_interspeech} and improve reliability \cite{parikh-etal-2026-rubric}.
For unbalanced data, Simple Preference Optimization (SimPO) \cite{NEURIPS2024_e099c1c9} better facilitates extending models toward multi-aspect or multi-granular assessment \cite{wang2025fine}.
Yet, E2E rubric-guided L2 comprehensive assessment that also 
generates label-consistent rationales remains relatively underexplored.

Despite this progress, instruction-tuned SpeechLLMs for reliable L2 speech assessment face three key gaps. First, a comprehensive multi-aspect and multi-granular assessment is rarely addressed in a single E2E setting: existing systems typically focus on sentence-level proficiency scoring or on localized word/phoneme diagnostics, but not both jointly. Second, training reliable models for ordinal scoring and fine-grained labeling is limited by severe label imbalance in L2 assessment data, with a strong skew toward correct productions and higher-rated categories compared to lower-rated ones. Third, while SpeechLLMs can generate free-text rationales alongside labels, the reliability of such rationales is often unclear: they may be superficially plausible yet insufficiently faithful to the assigned labels. In practice, this requires rationales to be (i) self-consistent with the model's predicted aspect labels (aspect mentions should be reflected in the corresponding predicted scores), and (ii) aligned with human judgments at both coarse polarity and finer aspect- or token-level references \cite{agarwal2024faithfulness}. 

To address these gaps, we propose a novel E2E SpeechLLM framework that jointly predicts multi-aspect, multi-granular L2 assessment labels and generates a natural-language rationale in a single response. We combine SFT with preference-based optimization to better align generated outputs with rubric-consistent behavior. Unlike token-level cross-entropy, preference optimization uses pairwise comparisons between chosen (rubric-consistent) and rejected (rubric-inconsistent) responses, encouraging label--rationale consistency and improving learning from rare error cases under severe class imbalance. Since preference alignment can be sensitive under skewed supervision and adjacent ordinal ratings, we adopt Bounded Direct Preference Optimization (BDPO) \cite{cho-etal-2025-rethinking}, a bounded variant of Direct Preference Optimization (DPO) \cite{10.5555/3666122.3668460}, which limits the influence of the rejected response.
To the best of our knowledge, the joint generation of multi-aspect, multi-granular L2 labels together with label-consistent rationales has not been systematically studied. The above considerations motivate our central research question: \textit{To what extent can an E2E SpeechLLM jointly predict multi-aspect, multi-granular L2 labels and generate rationales consistent with its predictions and human judgments?}

\section{Methodology}

\subsection{Model Architecture}

We use the state-of-the-art (SOTA) open source Qwen2-Audio-7B-Instruct \cite{chu2023qwen} as the backbone SpeechLLM. To minimize memory footprint and preserve pretrained acoustic representations, the base model weights are frozen under 4-bit quantization \cite{10.5555/3666122.3666563}. Fine-tuning is performed using Low-Rank Adaptation (LoRA) \cite{hu2022lora}, injected into the query, value, output, up-projection, and down-projection modules of the decoder layers. We set the LoRA rank to $r=64$, scaling factor to $\alpha=128$, and dropout to $0.05$. This configuration results in approximately $115$M trainable parameters ($\sim 1.6\%$ of the original 7B model)\footnote{\url{https://github.com/Aditya3107/speechllm-l2-assessment}}.

\subsection{Hybrid Optimization Objective}

We train the model using a hybrid objective that combines SFT with preference-based alignment. We first define the preference tuples used for alignment, then present the DPO objective and motivate our use of its bounded variant, BDPO. We conclude with the final combined loss used in all experiments.

For preference optimization, two alternative label outputs for the same input are required: one chosen and one rejected. Given an input $x$ consisting of the speech file and rubric prompt, we construct preference tuples $(x, y_c, y_r)$, where $y_c$ is the chosen (ground-truth) set of ordinal labels and $y_r$ is a rejected set of ordinal labels (see Section~\ref{rejected_pair_construction}). Here, $y$ refers to the structured label output spanning sentence-level (Accuracy, Fluency, Prosody) and word- and phoneme-level Accuracy labels. The preference dataset is $\mathcal{D} = \{(x^{(i)}, y_c^{(i)}, y_r^{(i)})\}_{i=1}^{N}$, and DPO optimizes the model assigning higher likelihood to $y_c$ than to $y_r$.

DPO compares a trainable policy $\pi_{\theta}$ against a frozen reference policy $\pi_{\text{ref}}$; in our implementation, $\pi_{\text{ref}}$ is the same backbone model with LoRA adapters disabled. We define the log-likelihood ratio
\begin{equation}
r_{\theta}(y \mid x) = \log \pi_{\theta}(y \mid x) - \log \pi_{\text{ref}}(y \mid x),
\label{eq:ratio}
\end{equation}
and optimize the standard DPO objective
\begin{equation}
\mathcal{L}_{\text{DPO}} = -\mathbb{E}_{\mathcal{D}}
\left[
\log \sigma \left(
\beta \left[r_{\theta}(y_c \mid x) - r_{\theta}(y_r \mid x)\right]
\right)
\right],
\label{eq:dpo}
\end{equation}
where $\sigma(\cdot)$ is the sigmoid function and $\beta$ is a temperature hyperparameter that controls the strength of the implicit regularization toward $\pi_{\text{ref}}$.

In fine-grained speech assessment, the rejected label set $y_r$ is often a near miss rather than a categorically incorrect output (e.g., \textit{Good} vs.\ \textit{Excellent}). In such cases, DPO can induce overly aggressive updates by strongly down-weighting plausible alternatives. We therefore use Bounded Direct Preference Optimization (BDPO) \cite{cho-etal-2025-rethinking}, which retains the same $\beta$ but limits how strongly rejected labels can be down-weighted by replacing the rejected term with a smoothly bounded alternative:
\vspace{-0.2cm}
\begin{equation}
\mathcal{L}_{\text{BDPO}} = -\mathbb{E}_{\mathcal{D}}
\left[
\log \sigma \left(
\beta \left[r_{\theta}(y_c \mid x) - z(x,y_r)\right]
\right)
\right],
\label{eq:bdpo}
\end{equation}
where the bounded rejection term is defined as:

\vspace{-0.5cm}
\begin{equation}
z(x, y_r) =
\log\left(
(1-\delta)\exp\left(r_{\theta}(y_r \mid x)\right) + \delta
\right),
\label{eq:margin}
\end{equation}
and $\delta\in(0,1)$ bounds the rejected term by replacing $\pi_{\theta}(y_r\mid x)$ with its interpolation with the reference,
denoted $\pi_{\text{mix}}(y_r\mid x)=(1-\delta)\pi_{\theta}(y_r\mid x)+\delta\pi_{\text{ref}}(y_r\mid x)$,
thereby preventing $\pi_{\theta}(y_r\mid x)$ from being driven arbitrarily low under near-miss ordinal alternatives.
Finally, we combine SFT with preference-based alignment and optimize the hybrid objective:
\vspace{-0.2cm}
\begin{equation}
\mathcal{L}_{\text{total}} = \mathcal{L}_{\text{BDPO}} + \lambda \cdot \mathcal{L}_{\text{SFT}},
\label{eq:hybrid}
\end{equation}
where $\lambda$ balances format learning via $\mathcal{L}_{\text{SFT}}$ and rubric alignment via $\mathcal{L}_{\text{BDPO}}$. In all experiments, we set $\lambda = 1.0$. In our implementation, $\mathcal{L}_{\text{SFT}}$ is computed on the target assessment response using teacher forcing, with prompt tokens masked such that cross-entropy is applied exclusively to the assessment text.

\subsection{Dataset}
\label{Dataset}
We use the publicly available SpeechOcean762 (SO762) dataset~\cite{zhang21x_interspeech}, a widely used benchmark for pronunciation and speaking assessment. It contains 5000 English read-speech utterances (2500 train / 2500 test); we train on the official training split and evaluate on the held-out test split. SO762 includes child and adult Mandarin L1 speakers and provides sentence-, word-, and phoneme-level human annotations. We evaluate five dimensions: sentence-level accuracy, fluency, and prosody, and word- and phoneme-level accuracy. We exclude completeness due to its unclear definition and word stress because it is a binary signal incompatible with the ordinal rubric used for the other dimensions. 
SO762 uses a 0--10 scale for sentence/word scores and a 0--2 scale for phoneme scores. We discretize both scales into five ordinal labels (\textit{Excellent}, \textit{Good}, \textit{Average}, \textit{Bad}, \textit{Worst}). The training data are severely skewed: over 80\% of labels are \textit{Good} or \textit{Excellent}, leaving the lowest categories virtually unrepresented. 

\subsubsection{Rejected Pair Construction}
\label{rejected_pair_construction}
SO762 provides one ground-truth label set per utterance, so we synthesize the rejected label set $y_r$ by perturbing the chosen labels $y_c$. For sentence-level metrics, we perturb \emph{exactly one} aspect label (e.g., \textit{Excellent}$\rightarrow$\textit{Good}) to isolate single-attribute errors. For sequence-level tasks (word and phoneme accuracy), we perturb every label in the sequence while preserving token alignment and output structure. To counter the ``niceness bias'' (systematic over-estimation) in generative assessment, perturbations are asymmetric: $\sim$88\% of labels are downgraded and 12\% upgraded. Across the 2500 training utterances, the perturbed target is distributed roughly uniformly across all dimensions ($\sim$500 per dimension) to encourage balanced learning.

\subsection{Prompting Strategy}
To support multi-aspect, multi-granular assessment in a single pass, we use a single rubric-driven prompt (Figure~\ref{fig:prompt}) that elicits all five dimensions jointly. This design reduces prompt-to-prompt variability and encourages label consistency across aspects and granularities. We adopt the grading rubrics from Table~1 in SO762~\cite{zhang21x_interspeech}. For each utterance, the prompt combines the speech, orthographic transcript, target phoneme sequence, and rubric definitions, instructing the model to assign the five ordinal labels defined in Section~\ref{Dataset}.

Because the model generates free-form text, we enforce a fixed response format to ensure reliable extraction of the predicted labels (i.e., deterministic parsing) \cite{wang2025speechllm}: one label per sentence-level aspect, followed by aligned word and phoneme label sequences (e.g., \texttt{WORD/Label}). The response ends with a free-text \texttt{Rationale} that justifies the predicted labels. Finally, ordinal labels are parsed and mapped to integer IDs (Section~\ref{eval_protocol}) to compute continuous evaluation metrics.

\begin{figure}[h]
\centering
\begin{tcolorbox}[colback=gray!5,colframe=gray!50,boxrule=0.5pt,arc=2pt,left=2pt,right=2pt,top=2pt,bottom=2pt,title=\small \textbf{System Prompt Template},before upper=\raggedright]
\scriptsize
\textbf{Task:} Assess the speech comprehensively based on the following rubrics.

\textbf{1. Sentence Metrics:} \\
\textbf{Accuracy:} [Rubrics for \textit{Excellent}/\textit{Good}/\textit{Average}/\textit{Bad}/\textit{Worst}] \\
\textbf{Fluency:} [Rubrics for \textit{Excellent}/\textit{Good}/\textit{Average}/\textit{Bad}/\textit{Worst}] \\
\textbf{Prosody:} [Rubrics for \textit{Excellent}/\textit{Good}/\textit{Average}/\textit{Bad}/\textit{Worst}]

\textbf{2. Word Accuracy:} Rate each word inline (\textit{Excellent}/\textit{Good}/\textit{Average}/\textit{Bad}/\textit{Worst}) based on the rubrics. \\
\textbf{3. Phoneme Accuracy:} Rate each phoneme inline (\textit{Excellent}/\textit{Good}/\textit{Average}/\textit{Bad}/\textit{Worst}) based on the rubrics. \\
\textbf{Input:} \
Transcript: "{transcript}". \
Target Phonemes: "{phoneme sequence}". \

\textbf{Output Format:} \\
Accuracy: [Label] \\
Fluency: [Label] \\
Prosody: [Label] \\
Words: WORD1/Label  WORD2/Label \dots \\
Phonemes: p1/Label p2/Label \dots \\
Rationale: [Free-text justification]
\end{tcolorbox}
\vspace{-0.4cm}
\caption{Abridged prompt structure (full rubrics omitted for brevity). The model predicts multi-granular labels and a natural-language rationale from speech, transcript, and target phonemes.}
\label{fig:prompt}
\end{figure}

\subsection{Training Procedure}
We fine-tune the model on a single NVIDIA RTX A6000 (48GB) GPU using the AdamW optimizer. To reduce gradient instability often observed in preference-based fine-tuning \cite{10.5555/3666122.3668460}, we use a constant learning rate of $5\times10^{-6}$ without a scheduler. Due to memory constraints, we use an effective batch size of 16 via gradient accumulation (batch size 1 with 16 accumulation steps). For BDPO, we set the preference strength to $\beta=0.1$ and the margin bound to $\delta=0.5$ based on empirical tuning. We train for 14 epochs and monitor the training loss across granularities: sentence-level losses plateau around Epoch~7, while word- and phoneme-level training losses continue to decrease until Epoch~11. Since fine-grained performance is critical, we report results using the Epoch~11 checkpoint.

\subsection{Evaluation Protocol}
\label{eval_protocol}
Following standard protocols \cite{zhang21x_interspeech}, we map label predictions to numeric scores to compute the Pearson Correlation Coefficient (PCC) and Root Mean Squared Error (RMSE) on the SO762 test set. We report the multi-class Matthews Correlation Coefficient (MCC) to account for ordinal class imbalance. For PCC, predicted ordinal labels are mapped to the midpoint of their corresponding rubric score intervals.

We evaluate the generated rationale against two reference signals: the model's predicted (Pred) labels  (\textbf{Internal (Pred)}) and ground-truth (GT) human annotations (\textbf{External (GT)}). Motivated by the plausibility--faithfulness distinction in model-generated rationales \cite{agarwal2024faithfulness}, we report two analyses. First, we assess plausibility via sentiment consistency by classifying each rationale as Positive/Neutral/Negative and comparing it to the polarity derived from sentence-level accuracy under both references. Second, we assess faithfulness via mention-based agreement by extracting sentence-level aspect mentions (accuracy/fluency/prosody) and any referenced words/phonemes, and measuring agreement with the corresponding Internal (Pred) and External (GT) labels on the mentioned subset.\footnote{Sentiment classification and mention extraction were performed using \texttt{Qwen/Qwen2.5-7B-Instruct}.}

\section{Results and Discussion}

We evaluate our fine-tuned SpeechLLM on the SO762 test set. To validate the proposed multi-aspect, multi-granular assessment setting, we structure the analysis into three parts: (i) overall performance across sentence-, word-, and phoneme-level predictions; (ii) comparison against single-granularity models and prior SOTA systems; and (iii) reliability of generated rationales, including plausibility and faithfulness analyses.

\subsection{Multi-Granular Assessment Performance}

Table~\ref{tab:multi_granular_results} presents the overall performance of our unified multi-granular model. Performance is strongest at the sentence level, with the highest correlation on Fluency (PCC 0.73), followed by Prosody, indicating effective modeling of broader suprasegmental cues (e.g., rhythm and intonation).

Correlation decreases on finer-grained sequence tasks (Word and Phoneme Accuracy), reflecting the known difficulty of token-level assessment. However, Word Accuracy achieves a higher MCC than Sentence Accuracy. This indicates that while continuous correlation (PCC) drops at the micro-level, the model remains effective at categorizing localized word-level errors. Finally, the lower RMSE for Phoneme Accuracy is due to its distinct 0--2 scoring scale, compared to the 0--10 scale used for all other metrics.



\begin{table}[ht!]
\centering
\caption{Performance of Multi-Granular SpeechLLM} 
\vspace{-0.2cm}
\label{tab:multi_granular_results}
\setlength{\tabcolsep}{4pt}      
\renewcommand{\arraystretch}{1.05} 
\footnotesize 
\begin{tabular}{@{}llcccc@{}}
\toprule
\textbf{Level} & \textbf{Task} & \textbf{PCC $\uparrow$} & \textbf{RMSE $\downarrow$} & \textbf{MCC $\uparrow$} \\
\midrule
\multirow{3}{*}{Sentence} & Accuracy & 0.66 & 1.72 & 0.35  \\
                          & Fluency  & 0.73 & 1.33 & 0.47  \\
                          & Prosody  & 0.71 & 1.48 & 0.42  \\
\midrule
Word    & Accuracy & 0.52 & 1.75 & 0.39  \\
Phoneme & Accuracy & 0.42 & 0.36 & 0.31  \\
\bottomrule
\end{tabular}
\end{table}

\subsection{Multi-Granularity vs. Single-Granularity Assessment}

To validate our approach, we compare the multi-granular model (Table~\ref{tab:multi_granular_results}) against single-granularity models fine-tuned independently for each granularity (Table~\ref{tab:single_granular_results}). All models use identical settings, differing only in restricted prompt structures.

The multi-granular model improves Sentence Accuracy across all metrics while maintaining comparable performance on Fluency and Prosody. We hypothesize this gain stems from the explicit inclusion of fine-grained phonetic and lexical targets during joint training, which helps ground the model's sentence-level judgments.

At the sequence level, the single-granularity model achieves a higher PCC for Word Accuracy, but the multi-granular approach achieves better MCC. This suggests that holistic sentence context helps the model separate distinct word-level error categories. For phonemes, multi-granular training improves PCC and RMSE but reduces MCC, indicating a trade-off between score calibration and strict boundary discrimination. Overall, the multi-granular model successfully balances holistic and fine-grained assessment without substantial degradation, supporting our joint multi-granular prompting strategy.



\begin{table}[ht!]
\centering
\caption{Performance of Single Granularity SpeechLLM}
\vspace{-0.2cm}
\label{tab:single_granular_results}
\setlength{\tabcolsep}{4pt}      
\renewcommand{\arraystretch}{1.05} 
\footnotesize
\begin{tabular}{@{}llcccc@{}}
\toprule
\textbf{Level} & \textbf{Task} & \textbf{PCC $\uparrow$} & \textbf{RMSE $\downarrow$} & \textbf{MCC $\uparrow$} \\
\midrule
\multirow{3}{*}{Sentence} & Accuracy & 0.62 & 1.84 & 0.34  \\
                          & Fluency  & 0.72 & 1.34 & 0.48  \\
                          & Prosody  & 0.71 & 1.44 & 0.46  \\
\midrule
Word    & Accuracy & 0.57 & 1.73 & 0.36  \\
Phoneme & Accuracy & 0.40 & 0.42 & 0.44  \\
\bottomrule
\end{tabular}
\end{table}

\subsection{Comparison with SOTA Approaches}

Table~\ref{tab:pcc_comparison_baselines} compares our single-granularity (BDPO-S) and multi-granularity (BDPO-M) SpeechLLMs with prior approaches, including the Goodness of Pronunciation Transformer (GOPT) \cite{9746743}, Azure Pronunciation Assessment (Azure PA) \cite{wang2025exploring}, and SimPO \cite{wang2025fine}. BDPO-M is competitive on sentence-level tasks and outperforms the LLM-based SimPO on fine-grained tasks, with higher PCC on Word and Phoneme Accuracy.

GOP-based ASR approaches, such as GOPT, still perform better on phoneme-level scoring, highlighting a remaining gap for end-to-end SpeechLLMs. However, BDPO-M also generates a natural-language rationale alongside proficiency labels, resulting in a more interpretable assessment output.

\begin{table}[ht!]
\centering
\caption{PCC scores using the SO762 dataset}
\vspace{-0.2cm}
\label{tab:pcc_comparison_baselines}
\setlength{\tabcolsep}{2pt}
\renewcommand{\arraystretch}{1.0}
\scriptsize
\begin{tabular}{@{}llccccc@{}}
\toprule
\textbf{Level} & \textbf{Task} & \textbf{GOPT \cite{9746743}} & \textbf{Azure PA \cite{wang2025exploring}} & \textbf{SimPO \cite{wang2025fine}} & \textbf{BDPO-S} & \textbf{BDPO-M} \\
\midrule
\multirow{3}{*}{Sentence} 
 & Accuracy  & 0.71 & 0.70 & 0.68 & 0.62 & 0.66 \\
 & Fluency  & 0.75 & 0.72 & 0.73 & 0.72 & 0.73 \\
 & Prosody  & 0.76 & 0.84 & 0.73 & 0.71 & 0.71 \\
\midrule
Word    & Accuracy & 0.53 & 0.62 & 0.51 & 0.57 & 0.52 \\
Phoneme   & Accuracy & 0.61 & --   & 0.38 & 0.40 & 0.42 \\
\bottomrule
\end{tabular}
\end{table}

\subsection{Plausibility and Faithfulness of Rationales}
Following the protocol described in Section~\ref{eval_protocol}, Table~\ref{tab:sentiment_alignment} reports the sentiment consistency of the generated rationales. Against the Internal (Pred) labels, the multi-granular model demonstrates high self-consistency: cross-polarity mismatches (e.g., Positive label with Negative rationale sentiment) are virtually absent, and most mismatches are one-step softenings (e.g., Negative$\rightarrow$Neutral) rather than reversals. However, relative to the External (GT) labels, a positivity bias emerges for lower-proficiency speech: the model frequently produces neutral-toned rationales even when human annotators assign negative labels, suggesting a systematic softening tendency.

\begin{table}[ht!]
\centering
\caption{Sentiment alignment matrix ($N=1941$; subset with successfully parsed rationales and sentiment labels). Cell values are row-wise percentages (\%) reported as \textit{Internal (Pred) / External (GT)}}
\vspace{-0.2cm}
\label{tab:sentiment_alignment}
\footnotesize 
\begin{tabular*}{\columnwidth}{@{\extracolsep{\fill}}lccc@{}} 
\toprule
\multirow{2}{*}{\begin{tabular}{@{}l@{}}\textbf{Polarity} \\ \textbf{(Pred / GT)}\end{tabular}} & \multicolumn{3}{c}{\textbf{Model Explanation Sentiment}} \\
\cmidrule(l){2-4}
 & \textbf{Negative} & \textbf{Neutral} & \textbf{Positive} \\
\midrule
\textbf{Negative} & \textbf{88.1 / 44.5} & 11.9 / 47.4 & 0.0 / 8.0 \\
\textbf{Neutral}  & 10.8 / 26.3 & \textbf{89.2 / 52.6} & 0.0 / 21.1 \\
\textbf{Positive} & 0.0 / 2.1 & 8.9 / 18.0 & \textbf{91.1 / 79.9} \\
\bottomrule
\end{tabular*}
\end{table}

Finally, Table~\ref{tab:explanation_consistency_gt} evaluates faithfulness via mention-based agreement by comparing aspect and token references explicitly mentioned in the rationale against both reference labels. The multi-granular model shows strong sentence-level agreement with Internal (Pred), but faithfulness degrades at finer granularities. Token-level references are sparse, especially for phonemes, and show weak agreement with both Internal (Pred) and External (GT) labels when they occur. Qualitatively, the model tends to produce broad rationales for high-quality speech, while attempting more specific word- and phoneme-level justifications for lower-rated speech. However, these localized rationales often fail to reliably localize errors and may reflect text-based or orthographic heuristics (e.g., \textit{``The word TROOPS has a slight error in the second letter''}). Overall, the rationales are strongly self-consistent at the sentence level, but faithfulness degrades sharply for token-level diagnosis.

\begin{table}[ht!]
\centering
\caption{PCC Agreement of Rationale Mentions with Model Predictions (Internal) and Ground-Truth Labels (External)}
\vspace{-0.2cm}
\label{tab:explanation_consistency_gt}
\setlength{\tabcolsep}{2.6pt}
\renewcommand{\arraystretch}{1.05}
\footnotesize 
\begin{tabular}{@{}llccc@{}}
\toprule
\textbf{Level} & \textbf{Task} & \textbf{$N$} & \textbf{Internal (Pred)} & \textbf{External (GT)} \\
\midrule
\multirow{3}{*}{Sentence} 
 & Accuracy & 1826  & 0.87  & 0.61 \\
 & Fluency  & 1940  & 0.86  & 0.66 \\
 & Prosody  & 1905 & 0.84 & 0.63 \\
\midrule
Word (Specific) 
 & Accuracy & 256 & 0.50 & 0.35 \\
Phoneme (Specific) 
 & Accuracy & 173 & 0.20  & 0.07 \\
\bottomrule
\end{tabular}
\end{table}

\section{Conclusion}

In this work, we presented an E2E rubric-guided SpeechLLM for comprehensive L2 speech assessment. By combining SFT with BDPO on the Qwen2-Audio-7B-Instruct model, it jointly predicted rubric-aligned proficiency labels across sentence, word, and phoneme levels and generated a natural-language rationale in a single response. 
Our results demonstrated that this joint modeling approach balances holistic and fine-grained assessment, outperforming recent LLM base approach on sequence-level scoring and matching single-granularity models without substantial degradation. We also proved that BDPO improves rubric adherence and label–rationale consistency in generative L2 speech assessment, particularly under skewed rating distributions.

Our analysis of generated rationales showed strong sentence-level reliability and sentiment consistency, demonstrating that the model produces largely plausible rationales for its predicted labels. However, faithfulness degrades at finer granularities: token-level references are sparse and, when present, may reflect orthographic heuristics rather than reliably localizing speech errors. 
Overall, we proved that SpeechLLMs demonstrate greater reliability for holistic assessment and high-level diagnostic support than for precise token-level error localization. Because mention extraction was automated to enable scalable evaluation, the resulting fine-grained faithfulness scores should be interpreted as approximate rather than exact measurements. Advancing fine-grained faithfulness estimation and robust error localization is, therefore, a key direction to strengthen pedagogically useful generative speech assessment systems.

\section{Acknowledgments}
This publication is part of the project Responsible AI for Voice Diagnostics (RAIVD) with file number NGF.1607.22.013 of the research programme NGF AiNed Fellowship Grants which is financed by the Dutch Research Council (NWO).

\section{Generative AI Use Disclosure}
Generative AI tools were used for language editing and polishing, including grammar and phrasing. All scientific content, experimental design, analyses, results, and conclusions were developed, verified, and approved by the authors. The authors take full responsibility for the content of this paper, and no generative AI tool is listed as a co-author.

\bibliographystyle{IEEEtran}
\bibliography{mybib}

\end{document}